%% file: bmvc_final.tex
\documentclass{bmvc2k}

\usepackage{wrapfig}

\usepackage{graphicx}
\usepackage{todonotes}
\usepackage{amsmath}
\usepackage{amssymb}
\newcommand{\f}{\mkern-2mu f\mkern-3mu}
\usepackage{bm}
\usepackage{multirow}  
\usepackage{booktabs}
\usepackage{tabularx}
\definecolor{mygray}{gray}{.92}
\usepackage{subfigure}
\usepackage[rightcaption]{sidecap}
\usepackage{colortbl}

\usepackage{rotating}
\usepackage{comment}

\usepackage{eucal}
\DeclareSymbolFontAlphabet{\mathrm}    {operators}
\DeclareSymbolFontAlphabet{\mathnormal}{letters}
\DeclareSymbolFontAlphabet{\mathcal}   {symbols}
\DeclareMathAlphabet      {\mathbf}{OT1}{cmr}{bx}{n}
\DeclareMathAlphabet      {\mathsf}{OT1}{cmss}{m}{n}
\DeclareMathAlphabet      {\mathit}{OT1}{cmr}{m}{it}
\DeclareMathAlphabet      {\mathtt}{OT1}{cmtt}{m}{n}
\DeclareSymbolFont{operators}   {OT1}{cmr} {m}{n}
\DeclareSymbolFont{letters}     {OML}{cmm} {m}{it}
\DeclareSymbolFont{symbols}     {OMS}{cmsy}{m}{n}
\usepackage{makecell}
\usepackage{array}
\usepackage{enumitem}
\makeatletter
\def\thickhline{%
  \noalign{\ifnum0=`}\fi\hrule \@height \thickarrayrulewidth \futurelet
   \reserved@a\@xthickhline}
\def\@xthickhline{\ifx\reserved@a\thickhline
               \vskip\doublerulesep
               \vskip-\thickarrayrulewidth
             \fi
      \ifnum0=`{\fi}}
\makeatother

\newlength{\thickarrayrulewidth}
\title{Face Pyramid Vision Transformer}

\addauthor{Khawar Islam}{https://khawar-islam.github.io}{1}
\addauthor{Muhammad Zaigham Zaheer}{https://zaighamz.com}{2,4}
\addauthor{Arif Mahmood}{arif.mahmood@itu.edu.pk}{3}

\addinstitution{
 FloppyDisk.AI, Pakistan
}
\addinstitution{
 Mohamed bin Zayed University of \\ 
 Artificial Intelligence, Abu Dhabi, UAE
}
\addinstitution{
 Information Technology University, \\ 
 Pakistan
}

\addinstitution{
 Electronics and Telecommunications Research Institute, South Korea
}

\runninghead{k. islam, M. zaigham zaheer, A. Mahmood}{FPVT}

\begin{document}

\maketitle
%\vspace{-5mm}
\begin{abstract}

%Vision Transformers (ViTs) have shown impressive results on various computer vision tasks however,  training under limited computational resources is still not well addressed. Also, training schemes for ViTs  using limited training data are not well explored.  %we first conduct an empirical study on pure ViTs and Convolutional ViTs with ablation in several aspects. Secondly, 

A novel Face Pyramid Vision Transformer (FPVT) is proposed to learn a discriminative multi-scale facial representations for  face recognition and verification. In FPVT, Face Spatial Reduction Attention (FSRA) and  Dimensionality Reduction (FDR) layers are employed to make the feature maps compact, thus reducing the  computations.  An Improved Patch Embedding (IPE) algorithm is  proposed to exploit the benefits of CNNs in ViTs (e.g., shared weights, local context, and receptive fields) to model lower-level edges to higher-level semantic primitives.  Within FPVT framework, a Convolutional Feed-Forward Network (CFFN) is proposed that extracts locality information to learn low level facial information.  The proposed FPVT is evaluated on seven benchmark datasets and compared with ten existing state-of-the-art methods, including CNNs, pure ViTs, and Convolutional ViTs. Despite fewer parameters, FPVT has demonstrated excellent performance over the compared methods. Project page is available at \url{https://khawar-islam.github.io/fpvt/}

%Vision Transformers (ViTs) have shown impressive results and achieved top-performing results on image classification tasks. However, training recipes for ViTs in Face Recognition (FR) domain are still suffering, and training under limited computational resources is still not addressed. In this paper, we first conduct an empirical study on pure ViTs and Convolutional ViTs with ablation in several aspects. Secondly, we propose Optimized Pyramid Vision Transformer (FPVT) to learn multi-scale face representations adaptively for general and age-invariant FR; Face-Spatial Reduction Attention (F-SRA) to mitigate the massive use of hardware resources and computations of large feature maps. Thirdly, FPVT proposes Overlap Patch Embedding (OPE) utilizes all benefits of CNNs (e.g., shared weights, local context, and receptive fields) to model low-level edges to higher-level semantic primitives. Lastly, FPVT introduces a Locality Feed-Forward (LFF) that extracts locality information to learn more about local representation as well as consider the long-range relationship. Finally, we validate FPVT on seven benchmark datasets and achieve superior performance with fewer parameters compared to pure ViTs and CNNs.
\end{abstract}

% LFW- unconstrained faces
% Age-DB -- age-estimation
% CP-LFW cross-age face verification
%SLLFW-- face verification with similar looking
% CFPFP face verification
% CA-LFW--> Cross-age face verification
% VGG2_FP frontal profile face pairs
%-------------------------------------------------------------------------
\section{Introduction}
\label{sec:intro}
Transformer models have achieved excellent performance on numerous natural language processing
tasks such as machine translation, question answering, and text classification. Later on, these models have also been successfully employed on many computer vision tasks, such as  object detection~\cite{li2022exploring}, scene
recognition ~\cite{wu2021p2t}, segmentation ~\cite{strudel2021segmenter}, and image super-resolution ~\cite{lu2021efficient}. Although ViTs are applicable to many computer vision tasks, it is challenging to directly adapt these to pixel-level dense predictions particularly required for object detection and image segmentation tasks. It is because output feature maps of transformers are single-scale and low-resolution. Moreover the computational complexity and memory overhead are quite high even for relatively smaller input image sizes. To handle these issues, Wang et al.~\cite{wang2021pyramid} have recently proposed Pyramid Vision Transformer (PVT). They introduced a pyramid structure to reduce the sequence length as the network deepens, resulting in significant reduction of the computational complexity. In the current work, we proposed Face Pyramid Vision Transformer (FPVT), which incorporates further complexity reduction, improved patching strategy, and a loss function more appropriate for face recognition (FR) and verification tasks.

FR task is more challenging than object recognition and image classification tasks due to subtle inter-person discriminative attributes and significant intra-person variations. ViTs have not yet been well explored for the task of FR despite the presence of large scale datasets.  Recently, Zhu et al.~\cite{zhu2021webface260m} proposed a dataset with $4$M identities and $260$M face images. However, training a ViT on a million-scale dataset takes significant time and requires extensive hardware resources. Our work in the current manuscript addresses this problem by employing PVT particularly for the application of FR and face verification. 

%Recently, face recognition (FR) models have achieved remarkable results and are considered a good authentication approach in various scenarios such as public safety, access control, monitoring persons through autonomous surveillance systems. The success behind FR models is the availability of large-scale training and well-labeled datasets. Recently, Zhu et al.~\cite{zhu2021webface260m} proposed a dataset with $4$M identities and $260$M face images. \ques{Similarly, the private face dataset collected by Canon in $2020$ had $18.8$M images containing $0.6$M different identities. \comme{mention reference, which dataset it this?}} A massive-scale training with Deep Convolutional Neural Networks (DCNNs) helps a lot to achieve excellent performance. However, training on a million-scale dataset takes significant time and requires extensive hardware resources. \par

%Despite the success of CNNs on large scale \cite{raghu2021vision}, the performance still varies and CNN pixel carries some important features for target output, which creates more burden in terms of representations and computational cost. Also, CNN utilizes a pixel array approach where every pixel denotes varying significance and this information is domain-specific. As well as, CNNs have less capability to capture global information and understand images. It only focuses on pixel-wise features and ignores the  structural dependency of features and understanding of images. \par

Learning rich multi-scale features is a useful task for various problems \cite{wang2021pyramid,liu2021swin, islam2021face}. As for FR, patches may have diverse poses, expressions, and shapes, which make it necessary to learn multi-scale representation. %However, it is yet to be discovered whether ViTs are better for FR tasks or not.
In this work, we hypothesize that our proposed FPVT can  capture long-range dependencies and distance-wise pixel relations, by constructing a hierarchical architecture and attention mechanism based on spatial reduction. Our proposed FPVT architecture consists of four stages that generate multi-scale features resulting in  less training data requirement, reduced computational resources, reduced number of parameters, and improved FR performance. 
\par
Our proposed FPVT  uses the advantages of CNNs, such as shared weights, local context, and receptive fields, while maintaining the benefits of ViTs, including attention, global context, and generalization. 
%The proposed FPVT architecture uses a hierarchical structure leveraging all benefits of CNNs and as well as advantages of ViTs. 
First, the transformer is divided into four blocks to create a pyramid structure. In the input, we employ an improved-patch approach to tokenize face images and expand the patch window such that it overlaps its surrounding patches. This allows FPVT to capture local facial continuity resulting in improved performance. Second, we introduce a depth-wise convolution in the feed-forward block of transformer to decrease the number of parameters while achieving better performance than PVT. 
%Also, our approach enables FPVT to capture spatial context and reduce parameters while keeping each separate. Third, 
To tackle the memory complexity in the recognition paradigm, we strategically embed the Facial Dimensionalty Reduction (FDR) layer~\cite{li2021virtual} in the training pipeline to minimize the time and hardware cost of our method.

%\comme{what is each in keeping each separate? What is OOM?}

% Answer of each separate: spatial context are useful features in face image, we handle both problems spatial and reduce parameters in our FPVT.

%Answer of OOM: Out of memory error, Its parameters are much greater than those of the feature extraction network. The dimensions of its output are also in the millions. The cost of the storage and calculation of the FC layer easily exceeds current GPU capabilities (leading to an out of memory error, OOM) and results in training failure (CVPR 2021)

The main contributions of the current work can be summarized as follows:
\begin{itemize}
    \item We present Face Pyramid Vision Transformer (FPVT) to learn multi-scale discriminative  features and reduce the computation of large feature maps while achieving superior accuracy. Face-Spatial Reduction Attention (F-SRA) is designed to reduce the number of parameters.
    \item We introduce Improved Patch Embedding (IPE) which utilizes all benefits of CNNs to model lower-level edges to higher-level semantic primitives.
    \item Additionally, FPVT introduces a CFFN that extracts locality information to learn more about local representations as well as consider a long-range relationships.
    \item Face Dimensionality Reduction (FDR) layer is introduced to make the facial feature map compact using a data dependent algorithm.
    \item Extensive experiments are performed on seven  benchmark datasets, including LFW, CA-LFW, CP-LFW, Age-DB, CFP-FF, CFP-FP, and VGG2-FP. Our FPVT has achieved excellent results compared with existing state of the art methods.
\end{itemize}
   
The rest of the paper is arranged as follows. Literature review is summarized in Section \ref{sec:relatedWork}, the proposed Face Pyramid Vision Transformer (FPVT) is described in Section \ref{sec:FPVT}, experiments/results in Section \ref{sec:result}, and conclusion and future directions follow in Section \ref{sec:future}.

\section{Related Work}
\label{sec:relatedWork}
%Recent developments in vision transformers (ViTs) have resulted in notable progress in Machine Learning. Several researchers have introduced ViTs for classification tasks in several fields, including autonomous medical data analysis, tabular data learning, etc. 

After the remarkable success of transformers in natural language processing (BERT\cite{devlin2018bert} or GPT \cite{radford2021learning}), Alexey et al. \cite{dosovitskiy2020image} introduced a transformer for computer vision tasks and obtained superior results compared to the conventional CNN models. In such models, to treat an image as a sentence, it is reshaped into two dimensional flattened patches. Afterwards, similar to the class tokens in BERT, learnable embeddings are added to the embedded patches. Finally, trainable positional embeddings are added on top of patch representations to preserve positional information. Notably, transformer architectures generally rely on self-attention mechanism without utilizing convolutional layers.
\par
ViT \cite{dosovitskiy2020image} models generally require massive amount of training images and, consequently, significantly huge computational cost, which bottlenecks their applicability. To eradicate this issue, a vision transformer architecture was proposed by Hugo et al. \cite{touvron2020training} that is trained on merely $1.2$M images. In this work, the original ViT architecture \cite{dosovitskiy2020image,masroor2019transtech} was modified to adapt teacher-student learning approach while enabling the native distillation process particularly designed for transformers. To this end, the output of the student network is learned from the output of the teacher network. Moreover, a distillation token was added to the transformer, which interacts with classification vectors and image component tokens. In convolution based models, we can generally enhance the performance by adding more convolutional layers. However, transformers are different in this sense and can quickly saturate if the architecture gets deeper. It is because the attention maps become less distinguished as we go deeper into the transformer layers. To eradicate this issue,
Daquan et al. \cite{zhou2021deepvit} proposed a re-attention mechanism, which regenerates the attention map to enhance the diversity between layers at a minute computational burden. This way, the re-attention module was successfully trained on a 32-layer architecture \cite{dosovitskiy2020image} to achieve 1.6\% improvement in the top-1 accuracy on Image-Net. Hugo et al. \cite{touvron2020training} focused on the optimization part in transformers and proposed CaiT which is similar to an encoder-decoder architecture.
Hugo et al. \cite{touvron2021going} also proposed to explicitly split class-attention layers dedicated to extracting the content of the processed patches from transformer layers responsible for self-attention among patches, which further enhances the performance.
\par
Existing transformers \cite{carion2020end,touvron2020training,yuan2021tokens,islam2020framework} have been highly focused on training ViTs from scratch and re-designed the token-to-token process which is helpful in modeling images based on global correlation and local-structure information. This also helps slightly in overcoming the need of deep and hidden layer dimensions. Chen et al. \cite{chen2021crossvit} also attempted to reduce the computational complexity of ViTs by introducing dual-branch ViT. The idea was to first extract multi-scale feature representations, then combining them using a cross-attention-based token-fusion mechanism. However, such models are still computationally expensive. Different from these prior works, our FPVT method is resource-efficient, thus works under limited computational resources. FPVT extracts local features while capturing global relationships with fewer parameters than ResNet-18 and recent ViTs.

\begin{figure}[t]
    \centering
    \includegraphics[width=\linewidth,keepaspectratio]{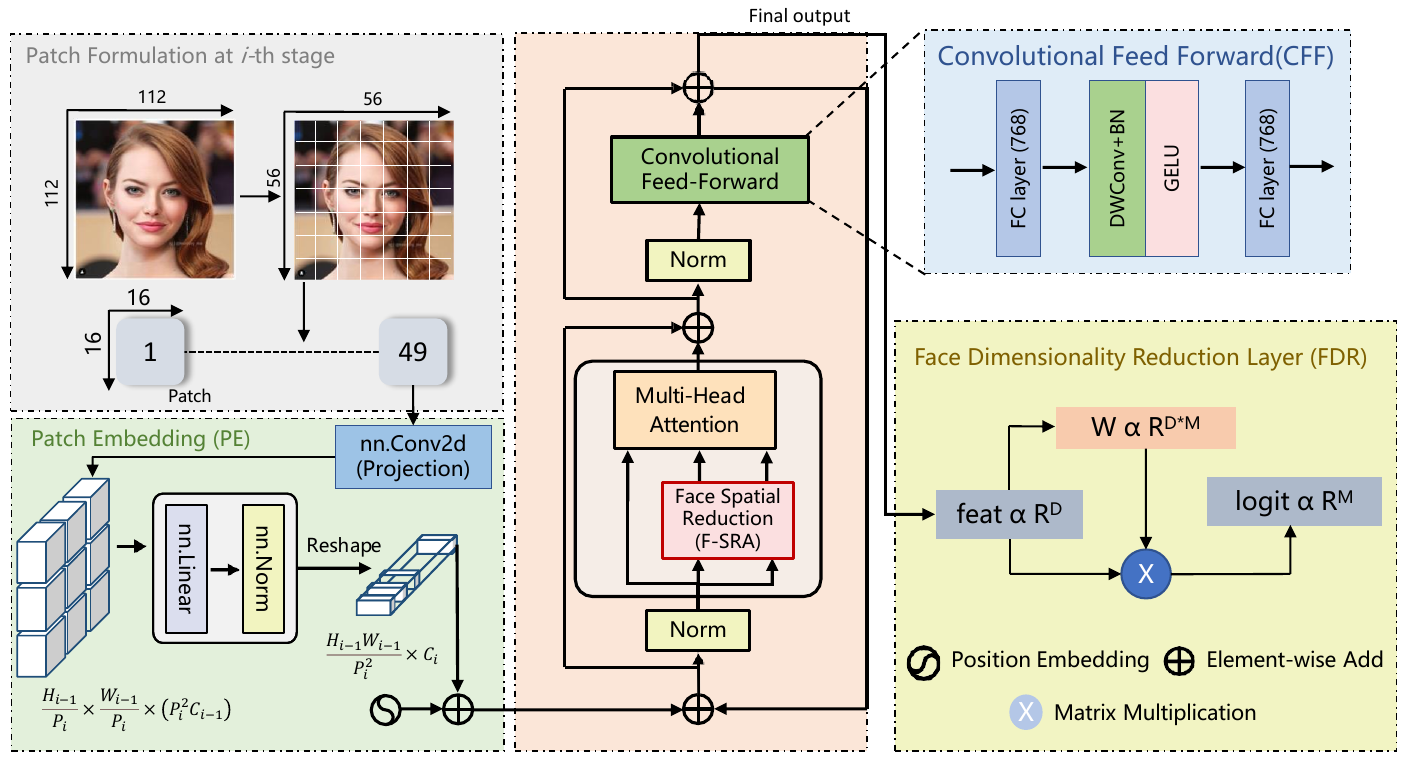}
    \caption{A simplified view of Face Pyramid Vision Transformer (FPVT) capable of training under limited computational resources. Each stage comprises of an improved patch embedding layer and an encoder layer. Following progressive shrinking strategy \cite{wang2021pyramid} , the output resolution is diversified at every stage from high to low resolution.}
    \label{fig:overall_architecture1}
\end{figure}
\section{Face Pyramid Vision Transformer}
\label{sec:FPVT}
\subsection{Overall Architecture}
We propose pyramid feature network, which has the capability to extract proportional sized features at high to low levels at different stages. The proposed \emph{Face Pyramid Vision Transformer (FPVT)} is a single network for general and age-invariant face recognition (FR). The network diagram of our FPVT is presented in Fig \ref{fig:overall_architecture1}. Similar to \cite{wang2021pyramid,liu2021swin}, our FPVT has four different pyramid stages that generate hierarchical feature maps. The construction of our FPVT comprises of improved patch embedding, face spatial reduction attention and convolutional feed-forward network. After that, face dimensionality reduction layer is responsible to compute discriminative compact facial features.
At the beginning of our method, given an input face image of size
$w \times h \times 3$, we split image into overlapping patches each of size $p_s \times p_s$ with overlap of $q_s \times q_s$ pixels which varies with the variation of stage $s$. The number of such patches turn out to be $(w/(p_s-q_s)-1)\times (h/(p_s-q_s)-1)$ for the stage $s$.   
We flatten these patches, feed to an improved patch embedding module, and get embedded patches of size $p_s^2\times c_s$, where $c_s$ is the number of channels at stage $s$. Next, positional embedding is attached with embedded patches and fed into encoder module. The output of the encoder is reshaped to be input to the next stage. The reshaped tensor has size $p_s^2 \times c_s$. The features obtained  from the  stage $s$ are fed into the next stage $s+1$. %In this way, we obtain four different feature maps: $\f_2$, $\f_3$, and $\f_4$, whose strides are $4$, $2$, $2$, and $2$. 
With pyramidal face features using $k$ stages, features $\{\f_1, \cdot \cdot \cdot \f_k\}$ are the output of each stage at a different resolution level. Such an approach is found to be suitable for general as well as age-invariant FR tasks.

%
%
\begin{comment}

\begin{wrapfigure}{R}{0.53\textwidth}
    \vspace{-10pt}
    \includegraphics[width=190pt, keepaspectratio]{images/difference.pdf}
    \caption{In CNNs, we give whole image to extract distinct features. In ViTs, image is split into patches to feed into transformer. In our approach, we suggest that the patch overlapping strategy helps to form a strong relationship between patches.}
    \label{fig:my_label}
\end{wrapfigure}

\end{comment}
\subsection{Improved Patch Embedding}
Instead of using non-overlapped patches from a face image, we use a simple yet effective technique to increase the performance of ViT's for FR in various scenarios. Motivated by recent work \cite{zhong2021face}, we introduce a token generation strategy in ViTs. We add a token generation scheme in the transformer \cite{wang2021pyramid} to generate sliding overlapped patches   and use inter-patch information to increase FR performance. 
\par 
In our Improved Patch Embedding (IPE), we utilize a convolution layer with padding $f$ to generate these patches. Overlap embedding enables FPVT to extract sequential information from faces while also reducing sequence length and increasing the feature dimension  over consecutive stages. It accomplishes spatial down-sampling while simultaneously increasing the number of feature maps. FPVT takes 2d input image from training data with size $h \times w \times c$, and feed into convolution layer with kernel size 2$f$+$1$, stride $s$,  number of kernels $p$, and padding size $f$. The final output is $\frac{h}{s} \times \frac{w}{s} \times p$. %Generally, the size of patches is chosen as defined in different ViTs, where a small patch size  gives a longer sequence of tokens. 
The IPE layer allows us to adjust the number of visual tokens and feature dimensions at each stage by using a convolution operation.

%\subsection{Locality Feed-Forward}
\subsection{Convolutional Feed-Forward Network}
When considering global relationships in visual recognition tasks, transformers are the first-priority to create  long-range dependencies via self-attention approach. However, transformers require significant computational cost. In order to reduce the computational complexity, we propose light-weight convolutional filters inspired by MobileNet architecture \cite{howard2017mobilenets}. %Different from the recent ViTs \cite{li2021localvit, liu2021swin, yuan2021tokens}, we propose to use MobileNet \cite{howard2017mobilenets} architecture 
These filters are helpful in capturing local features from a face image, e.g., forehand lines, nose pattern, nose bridge, and chin. Particularly, we introduce a set of $h=3 \times w=3$ filters having cardinality of $n_i$ and the number of input channels and padding of $1$. Then, a set of $1 \times 1$ depth-wise convolution and cardinalty $n_o$ is applied to conduct across channel convolutions. Our light weight filters require only $hwn_i+n_in_o$ parameters, which are significantly lesser than $hwn_in_o$ parameters in the equivalent normal convolution filters.

% with batch normalization layer only which avoids more less complexity compare to \cite{li2021localvit} and quite friendly for feed-forward networks. 
%We generate the $\mathcal{I}_{ftr}$ and $\mathcal{O}_{ftr}$ with dimensions $\mathcal{D}$ based on the $\mathcal{H}\times \mathcal{W}$ of a patch and number of channels $\mathcal{NC}$. 
\begin{comment}

\begin{wrapfigure}{L}{0.31\textwidth}
    \includegraphics[width=110pt, keepaspectratio]{images/depthwise.pdf}
    \caption{Transformer encoder layer with our convolutional feed-forward network.}
    \label{fig:my_label}
\end{wrapfigure}
\end{comment}

Our convolutional feed-forward network comprises of one fully connected layer, a light weight convolutional layer, a batch-wise normalization layer,  a GELU activation function, followed by another linear layer. Such an architecture brings rich representations and gives low-level information which is not addressed in the previous feed-forward networks. %In a view of FPVT complexity, our depth-wise convolution is more light-weight and comparable better than previous feed-forward networks in term of performance. 
The depth-wise convolution is obtained in two steps. 
First we convolve each channel $W_q \in \mathbb{R}^{h\times w}$ with a filter $Y_q \in \mathbb{R}^{m\times m}$ to get 
\begin{equation}
 D_q= W_q \odot Y_q, \text{ where } 1 \le q \le n_i,
\end{equation}
where $W_q \in \mathbb{R}^{h\times w}$  is the result of q-th channel convolution. In the second step, depth-wise convolution is performed by using a set of $1\times 1$ filters $\mathbf{v}_p \in \mathbb{R}^{n_i}$: 
\begin{equation}
F_{p}(a,b)= \sum_{q=1}^{n_i} D(a,b,q)\mathbf{v}_p(q), \text{ where } 1\le p \le n_o
\end{equation}
$F_p$ is the final output features of our light-weight convolutional feed-forward network. After that, the output features are reshaped to generate a sequence of tokens which are used to feed into the next transformer layer.
%Since only one depth-wise convolution with BatchNorm is applied to current features, it can capture local information among adjacent features. Note that, our approach is efficient to extract spatial features and generalize better to other well-performed architectures. In this way, compared to existing ViTs, the representation efficiency and parameters are decreased. The computational cost can be presented as:

\subsection{Face Spatial Reduction Attention (F-SRA)}

The proposed FPVT encoder at $i$-th stage has $l_{i}$ encoder layers and each stage consists of a convolutional feed-forward network and a self-attention \cite{vaswani2017attention}. Since FPVT requires to process low-resolution face images for constructing hierarchical feature maps, instead of utilizing standard Multi-Head Attention (MHA) layer, we introduce a simple yet effective Face Spatial-Reduction Attention (F-SRA) layer. Compared to MHA, our F-SRA requires three inputs including query $q$, key $k$, and value $v$ while the output consists of refined features. Before attention process, our F-SRA decreases spatial scale of $k$ and $v$. In this way, F-SRA has low  computational cost and reduced memory overhead. Our F-SRA at the $i$-th stage having $c_i$ channels and ${c_i}$ heads is given by the concatenation of all heads:

$     \text{SRA}(q, k, v) = [h_0, h_1, \cdots, h_j , \cdots h_{c_i}] w_o,
%    \label{eqn:att1}
$
where $w_o$ is a linear projection matrix, %$w^o\in\mathbb{R}^{n_i\times n_i}$  
and $h_j$ is the output  of $j-$th head:  
\begin{equation}
     h_j=\rm{Att}(q w_j^q,  s_r(k)w_j^k, s_r(v)w_j^v).
    \label{eqn:att2}
\end{equation}
 Where ${\rm Att}(q, k, v) = {\rm Softmax}(\frac{qk^\top}{\sqrt{d_{\rm h}}})v$, and the parameters for linear projection operation are $w_j^q \in \mathbb{R}^{c_i\times d_{ h}}$,
$w_j^k\!\in\!\mathbb{R}^{C_i\!\times\!d_{h}}$,
$w_j^v\!\in\!\mathbb{R}^{c_i\!\times\!d_{h}}$. Dimension of head, $d_{ h}$ is equal to $\frac{c_i}{n_i}$. ${s_r}(\cdot)$ is the technique for decreasing the spatial dimension of input sequence $k$ and $v$ %which can be represented as:
\begin{equation}
    {s_r}(x_i, r_i, w_s) = {\rm Norm}({ \rm Reshape}(x_i, r_i)w_s).
    \label{eqn:att4}
\end{equation}
where $x_i\!\in\!\mathbb{R}^{(h_iw_i)\!\times\!c_i}$ denotes an input sequence, and $r_i$ represents the reduction ratio of the attention layers at $ith$ stage. $\rm{Reshape}(x, r_i)$ is the method to reshape the input sequence $x_i$ to a sequence of size $\frac{h_iw_i}{r_i^2}\!\times\!(r_i^2c_i)$. $w_s\!\in\!\mathbb{R}^{(r_i^2c_i)\!\times\!c_i}$ is a linear projection that decreases the dimension of the input sequence to $c_i$. ${\rm Norm}(\cdot)$ represents layer normalization. Through the above mathematical representations, we can compute the memory-cost of attention operation which is $r_i^2$ smaller than MHA. Despite the fact that, attention mechanism have strong potential for learning global relationship, the computational overhead of feature maps is still expensive. Thus, we utilize adaptive max pooling layer with output size $7$ over an input feature before attention operation. This technique brings a substantial reduction in parameters and handles large feature maps with less computational resources. %The equation can be written as follows:
\begin{equation}
     Y = \text{F-SRA}(\rm{AdaptiveMaxPool}( \rm{SRA}(\mathbf{q}, \mathbf{k},\mathbf {v}))),
\end{equation}
This further reduces the size of input matrix by a factor of $n_{mp}^2$, where $n_{mp}\times n_{mp}$ is the size of max pooling filter. 
% ${\rm Att}(\cdot)$ can be presented as:
%\begin{equation}
%    {\rm Att}(q, k, v) = {\rm Softmax}(\frac{qk^T}{\sqrt{d_{\rm h}}})v
%    \label{eqn:att3}
%\end{equation}
%
%
%
%
\vspace{-2mm}
\subsection{Face Dimensionality Reduction Layer}
While FPVT extracts multi-scale features, we also require dimensionality reduction mechanism for training ultra large-scale dataset with limited hardware costs. Inspired by the recent advances in computationally efficient FR methods \cite{li2021virface, wang2021efficient}, we introduce a Face Dimensionality Reduction (FDR) layer in the ViT stream which reduces training time while also maintaining superior accuracy. \par
%\vspace{-12pt}
%\paragraph{Training Pipeline.} 
In the training phase, FDR layer randomly splits $k$ training identities (categories) into $m_g$ groups. The categories from $m_g$ share the $l_{th}$ column in projection matrix $w$. In this paper, $l_{th}$ column is defined as $anchor$ and $w$ contains anchors shared by $m_g$. To optimize $w$, we initialize two anchors, corresponding anchor $anch_{corr}$, and free anchor $anch_{free}$. If mini-batch carries categories from the $l_{th}$ column then $anch_{l}$ is of type $anch_{corr}$. If not, it is marked as $anch_{free}$.
\vspace{-15pt}
\paragraph{Corresponding Anchor.} If mini-batch carries a category from $m_g$, $anch_{l}$ is placed in $anch_{corr}$ in that iteration. With each column in $w$ of the last FC layer representing the centroid of each category, the equation of $anch_{corr}$ can be written as:

\begin{equation}
\label{equ:anchor}
anchor_{corr,l}={\sum_{i=1}^{K} \alpha_{i, l} f_{i, m}}/{\sum_{i}^{K} \alpha_{i, l}}
\end{equation}
$f\textsubscript{i, m}$ is feature representation of the $l_{th}$ face belong to $m_g$. We assume that there is no conflict among anchors. $\left\{f_{i, m}\right\}(i=1,2,3,4 \ldots, k)$ represents an individual identity. $\alpha_{i, l}$ is an estimated attention factor, $f_{i, m} \cdot\left\{\alpha_{i, l}\right\}$ is estimated through attention process or set as a constant value. 
\textbf{{Free Anchor:}} Due to the limited resources, it is not possible to set a larger batch size and $anch_{corr}$ is also restricted by batch size. To overcome this limitation, the concept of free anchors is introduced. If face does not exist in $m_g$ in a given iteration, $anch_{l}$ will be free and represented as $anch_{free, l}$. This way, it cannot be calculated by Equation \ref{equ:anchor} due to $f_{i,l} \in \theta$. The concept of free anchors help in ending the restriction of large batch number in the same way as traditional FC layers. Inter-identity representation can also be dispersed among mini-batches. Moreover, the number of $m_g$ could be set independently based on computational resources and accuracy. Free anchors are not restricted by the number of samples or batch size.
The FDR layer is comparably better than a traditional FC layer especially with limited hardware resources. The kernel of FDR layer is $w \in \mathbb{R}^{d \times m}$ where $m$ is the hyperparameter which depend on the balance between hardware resources and performance and can be set freely. The number of $m$ must be less than the training categories $n$. The final output of FDR layer is represented as: $y=w^{T} f+b$.
Here, $y \in \mathbb{R}^{m}$ is final output, $b$ is bias, and $f \in \mathbb{R}^{d}$ is feature. In the representation learning case, $b$ can be written as zero.

\vspace{-2mm}
%\subsection{Model Details}
%The proposed FPVT parameters are described as follows: For the $i_{th}$ stage, $p_i$ is the patch-size, $c_i$ is the number of output channel, $l_i$ is the number of layers in encoder, $r_i$ is the reduction-ratio in F-SRA, $h_n$ is the number of heads, $e_i$ is the expanding-ratio of convolutional FFN.

%Following the design principles of SwinT \cite{liu2021swin} and PyramidT \cite{wang2021pyramid}, we utilize the small number of output channels in shallow stages and focus  the major computational resource  in the middle stages. To provide instances of FPVT, we present only one model of our method which is presented in Table.~\ref{tab:arch4}. The  number of parameters of FPVT is smaller than ResNet-18 \cite{he2016deep}, IR-18 \cite{deng2019arcface}, IR-SE-18 \cite{hu2018squeeze}.

\vspace{-2mm}
\section{Experiments}
\label{sec:result}
We performed extensive experiments to evaluate our proposed FPVT %the robustness of CNNs, pure ViTs, , and our FPVT 
on several benchmark datasets and compared with CNN based methods \cite{he2016deep, hu2018squeeze, deng2019arcface}, pure ViT methods \cite{dosovitskiy2020image, zhou2021deepvit,zhang2021aggregating, touvron2021going}, and Convolutional ViTs \cite{heo2021rethinking, yuan2021incorporating, wu2021cvt}. %We also conducted an extensive ablation study of the proposed FPVT algorithm.
%
\begin{comment}

\begin{figure}[!t]
    \subfigure[Training loss curves]{\includegraphics[width=0.245\linewidth]{images/all_vits.pdf}}
    \subfigure[Accuracy curves]{\includegraphics[width=0.245\linewidth]{images/acc_vit.pdf}}
    \label{fig:ALLVITS}
    \subfigure[Training loss curves]{\includegraphics[width=0.245\linewidth]{images/ours_loss.pdf}}
    \subfigure[Accuracy curves]{\includegraphics[width=0.245\linewidth]{images/ours_acc.pdf}}
    \caption{Comparison of our module integrates in FPVT.}
    \label{fig:OPVT}
\end{figure}

\end{comment}

\vspace{-2mm}
\subsection{Implementation Details}
%We propose the patch size at the first stage to be $7 \times 7$ in our method, and other three stages have a patch size of $3 \times 3$.
Following previous ViT works \cite{zhong2021face}, we use adamW optimizer with initial LR $3e-4$ for all pure ViTs and ConViTs experiments. For CNN works \cite{hu2018squeeze}, we use SGD and set the LR to be $0.1$ with momentum set to $0.9$ and use weight decay $0.05$. We train FPVT along with all methods for 60 epochs on the face scrub dataset. We use one Nvidia Tesla V$100$ supporting a batch size of $496$ in all of our experiments. We use ArcFace as a classification head in CNNs and ViTs. We employ standard data augmentation techniques which include resizing, random crop, and random horizontal flip.
\noindent\textbf{Training Dataset:}  Face Scrub dataset contains $107,818$ images of $265$ male and $265$ female celebrities collected from different sources on the internet. For training, a cleaned and aligned version of the dataset is used which includes $91,712$ images of $263$ males and $263$ females. 
%
%The statistics of an original and cleaned dataset are presented in Table. \ref{tab:FACESCRUB}. In female celebrities, Adrianne Leon and Joanna Garcia are missing from the cleaned dataset. 
%\par
%\vspace{5pt}
%
\textbf{Testing Dataset:} Tests are conducted on various databases including  Age-DB \cite{moschoglou2017agedb} for age-invariant, LFW for unconstrained FR \cite{huang2008labeled}, 
%SL-LFW for similar-looking faces, \cite{deng2017fine} (Serious Mistake)
CFP-FP \cite{sengupta2016frontal} for frontal-profile, CP-LFW \cite{zheng2018cross} for a cross\-pose, CA-LFW \cite{zheng2017cross} for cross-age, and VGG2-FP is for frontal pose). 
\textbf{LFW} \cite{huang2008labeled} dataset exhibits natural pose variations, focus, lighting, resolution, make-up, occlusions, background, facial expression, age, gender, race, accessories, and photographic quality variations. It comprises 13,233 images of faces gathered from online websites. 
%Most face images are in color, but few images are gray-scale only. 
\textbf{CA-LFW} purely consists of $3,000$ positive face aging pairs. It has been split into $10$ distinct folds using the similar identities included in the LFW $10$ folds. This database consists of $4,025$ individuals with $2$, $3$ or $4$ face images for each identity. 
\textbf{CP-LFW} \cite{zheng2018cross} is developed to consider pose-related images in FR. It is an extended version of the LFW dataset which consists of $11,652$ face images, $3,968$ unique identities, and two to three images per person.
\textbf{Age-DB} is manually collected data in wild with noise-free labels. 
%Mostly captured images under uncontrolled environment, real-world %situations (i.e., different poses, noise, expressions, occlusions, %etc.). 
It contains $16,488$ images of various actress/actresses, writers, politicians and etc. It consists of $568$ unique identities, $29$ images per person and age ranges start from $1$ to $101$ years. 
%It contains $5,749$ images with $1,680$ ID's. Verification accuracy is measured on $6,000$ face pairs. 
%\textbf{FGLFW} also known as a Similar-looking LFW (SLLFW) dataset  consists of $3,000$ same face pairs within the actual LFW dataset. FGLFW dataset split into $10$ separate subsets of face pairs for cross-validation. Each separate subset comprises $300$ matched face pairs and $300$ negative face pairs.
\textbf{CFP} contains $7,000$ face images, $500$ unique identities and number of images per subject is $14$ \cite{sengupta2016frontal}.The database is split into $10$ subsets with a pairwise separate set of identities in each split. Every subset comprises $50$ individuals and $7,000$ pairs of faces for frontal-frontal and frontal-profile (CFP-FP) experiments. \textbf{VGG2-FP} is extracted from large-scale VGGFace2 \cite{cao2018vggface2} dataset consists of 3.31M images of $9131$ categories with large range of ethnicity age and pose. It is specially designed for frontal-profile faces and it consists of $10,000$ images of $300$ identities with different variation. 
%\vspace{-20pt}
%
% Table 2. LFW,CA-LFW Age-DB,CP-LFW , 
\begin{wraptable}{R}{0.50\textwidth}
\scalebox{0.70}{
    \begin{tabular}{p{0.3cm}|p{2.0cm}|p{0.6cm}<{\centering}p{0.6cm}<{\centering}p{0.8cm}<{\centering}|p{1.1cm}<{\centering}}
    \hline\thickhline
    \rowcolor{mygray} 
    \multicolumn{2}{c|}{}  & \multicolumn{3}{c|}{LFW $(family)$} & \\ 
    \cline{3-5} 
    \rowcolor{mygray}
    \multicolumn{2}{c|}{\multirow{-2}{*}{Methods}} & \multicolumn{1}{c}{LFW} & \multicolumn{1}{c}{CA} 
    & \multicolumn{1}{c|}{CP} & \multicolumn{1}{c}{\multirow{-2}{*}{Age-DB}} \\
    \hline\hline
    \multirow{3}{*}{ \rotatebox{90}{CNN}}  
    & ResNet-18\cite{he2016deep}       & 76.7  & 60.7 & 58.1   & 61.4    \\ 
    & IR-50~\cite{deng2019arcface}    & 91.7  & 78.1 & 68.9   & 73.4 \\ 
    & IR-SE-50~\cite{hu2018squeeze}  & 90.5  & 65.8 & 68.7   & 65.8 \\
    \hline \hline
    \multirow{5}{*}{ \rotatebox{90}{PureViT}} 
    & DeepViT \cite{zhou2021deepvit}                 & 75.5  & 62.6 & 57.1   & 59.7 \\
    & CaiT \cite{touvron2021going}                   & 83.4  & 71.5 & 57.5   & 62.2 \\
    & ViT \cite{dosovitskiy2020image}                & 81.9  & 67.7 & 58.9   & 61.4 \\
& ViT \cite{dosovitskiy2020image}+\textbf{IPE}   & 82.5  & 68.5 & 61.1   & 63.1 \\
\hline\hline
\multirow{9}{*}{ \rotatebox{90}{ConViT}}  
& PiT \cite{heo2021rethinking}       & 80.6  & 66.6 & 58.7   & 64.6    \\ 
& CvT \cite{wu2021cvt}               & 82.5  & 69.1 & 57.1   & 63.7 \\ 
& CeiT \cite{yuan2021incorporating}  & 84.8  & 72.6 & 60.1   & 65.8  \\
& PVT \cite{wang2021pyramid}         & 78.8  & 66.8 & 55.1   & 59.9 \\
& \textbf{ +IPE}                     & 82.9  & 70.1 & 59   & 65.6 \\
& \textbf{ +CFFN}                    & 86.7  & 72.9 & 62.1   & 68.9 \\
& \textbf{ +FDR}                     & 87.4  & 73.9 & 61.6   & 70.1 \\
& \textbf{ +OA}                      & 91.4  & 77.4 & 68.9   & 74.5 \\
& \textbf{FPVT}                      & \textbf{92.0}  & 77.0 & 67.8   & \textbf{75.0} \\
\hline\thickhline
\end{tabular}
}
\vspace{10pt}
\caption{Face verification accuracy on LFW, CA-LFW, CP-LFW and Age-DB: Comparison with FPVT, CNNs, PureViTs and Convolutional ViTs methods. 
% The top first three models represents the results of most representative CNN models, while mid block indicates the performance of pure ViTs architectures. Third block shows the results of Conv. ViTs models.
}
\label{tab:table3}
\vspace{-5pt}
\end{wraptable}
\vspace{-3mm}
\subsection{Quantitative and Qualitative Comparison} 

%
%
%
%To reduce complexity,  number of parameters require to be trained should be reduced.  However, the number of parameters in IR-18 \cite{deng2019arcface} and IR-SE-18 \cite{hu2018squeeze} are doubled of ResNet-18 \cite{he2016deep}. 

FPVT is compared with IR-18 and IR-SE-18   which are industrial benchmarks for the FR tasks. To compare FPVT with pure ViTs, we utilize  ViT \cite{dosovitskiy2020image}, DeepViT \cite{zhou2021deepvit}, and CaiT \cite{touvron2021going}. Further, we compare FPVT with three convolutional ViTs namely PiT \cite{heo2021rethinking}, CeiT \cite{yuan2021incorporating} and CvT \cite{wu2021cvt}. For a fair comparison, we evaluate all models using popular FR metric Face Verification Accuracy (FVF).
%
%, as in VPL \cite{deng2021variational}. Here, threshold scores are computed during the training phase and the output of the trained model is embedding size to calculate the similarity score.
%
%
%
Table. \ref{tab:table_2} shows FPVT outperforms existing methods including pure ViTs, ConViTs, and CNNs in terms of face verification accuracy, on three datasets. The higher VA indicates the capability of FPVT to verify general and age-invariant faces. As we can see, ConViTs outperformed the pure ViTs models and are  proven to be the robust ConViT models CeiT \cite{yuan2021incorporating} and CvT \cite{wu2021cvt}, which require a large number of parameters to produce superior results. 
In contrast, FPVT does not need further training data and the number of parameters is less than existing  models.
\vspace{-2mm}
\subsection{Ablation Study}
We conduct multiple ablation studies to validate the impact of our proposed work in our FPVT modules. Table. \ref{tab:table3} and Table \ref{tab:table_2} present the results of CNNs, pure ViTs, and Convolutional ViTs on seven datasets. PVT (referred to as baseline) is a standard pure pyramid transformer without convolution. We choose PVT as a baseline due to two main reasons: $i$) It generates multi-scale features. $ii$) The number of parameters is larger than ResNet18. The introduction of the IPE block leads to a gain of 1\% in terms of performance, highlighting the impact of convolutional tokens. IPE block improves the performance on LFW from 78.8\% to 82.9\%, CFP-FF from 75.2\% to 85.5\%, CFP-FP from 52.9\% to 65.6\%, Age-DB from 59.9\% to 65.6\%, CA-LFW from 66.8\% to 70.1\%, and CP-LFW from 55.1\% to 59\%. 
\begin{wraptable}{R}{0.70\textwidth}
    \centering
	\scalebox{0.70}{
		\begin{tabular}{p{0.3cm}|p{2.0cm}|p{1.0cm}<{\centering}|p{1.0cm}<{\centering}|p{1.0cm}<{\centering}|p{0.5cm}<{\centering}p{1.2cm}<{\centering}|p{1.5cm}<{\centering}}
			\hline\thickhline
			\rowcolor{mygray} 
			\multicolumn{2}{c|}{}  & & & & \multicolumn{2}{c|}{CFP $(family)$} & \\ 
			\cline{6-7} 
			\rowcolor{mygray}
			\multicolumn{2}{c|}{\multirow{-2}{*}{Methods}} &
			\multicolumn{1}{c|}{\multirow{-2}{*}{Dim}} &
			\multicolumn{1}{c|}{\multirow{-2}{*}{Depth}} &
			\multicolumn{1}{c}{\multirow{-2}{*}{Param}} & 
			\multicolumn{1}{|c}{FF} & {FP} & 
			\multicolumn{1}{c}{\multirow{-2}{*}{VGG2-FP}} \\
			\hline\hline
			\multirow{3}{*}{ \rotatebox{90}{CNN}}  
			  & ResNet-18\cite{he2016deep}                   & -            & -          & 30.7M          & 76.7          & 52.2 & 61.4          \\ 
			  & IR-50~\cite{deng2019arcface}                & -            & -          & 65.1M          & 91.7          & 74.2 & 73.4          \\ 
			  & IR-SE-50~\cite{hu2018squeeze}              & -            & -          & 65.5M          & 90.5          & 71.6 & 65.8          \\
			\hline \hline
			\multirow{5}{*}{ \rotatebox{90}{PureViT}} 
			  & DeepViT \cite{zhou2021deepvit}               & 512          & 6          & 11.6M          & 75.5          & 56.1 & 59.7          \\
			  & CaiT \cite{touvron2021going}                 & 512          & 3          & 7.8M           & 83.4          & 56.6 & 62.2          \\
			  & ViT \cite{dosovitskiy2020image}              & 512          & 6          & 17.8M          & 81.9          & 58.9 & 61.4          \\
			  & ViT \cite{dosovitskiy2020image}+IPE & 512          & 6          & 17.9M          & 82.5          & 60.6 & 63.1          \\
			\hline\hline
			\multirow{9}{*}{ \rotatebox{90}{ConViT}}  
			  & PiT \cite{heo2021rethinking}                 & 64           & 20         & 12.5M          & 80.6          & 57.2 & 64.6          \\ 
			  & CvT \cite{wu2021cvt}                         & 64           & 10         & 19.8M          & 82.5          & 56.4 & 63.7          \\ 
			  & CeiT \cite{yuan2021incorporating}            & 64           & 20         & 21.5M          & 84.8          & 59.1 & 65.8          \\
			  & PVT \cite{wang2021pyramid}                   & 512          & 18         & 32.2M          & 78.8          & 52.9 & 59.9          \\
			  & \textbf{ +IPE}                               & 512          & 6          & 33.3M          & 82.9          & 56.4 & 65.6          \\
			  & \textbf{ +CFFN}                               & 512          & 6          & 33.3M          & 86.7          & 61 & 68.9          \\
			  & \textbf{ +FDR}                               & 512          & 6          & 33.3M          & 87.4          & 61.5 & 70.1          \\
			  & \textbf{ +OA}                                & 512          & 6          & 33.3M          & 91.4          & 71.8 & 74.5          \\
			  & \textbf{FPVT}                                & \textbf{512} & \textbf{6} & \textbf{28.2M} & \textbf{92.0} & \textbf{73.3} & \textbf{75.0} \\
			\hline\thickhline
		\end{tabular}
	}
	\vspace{2pt}
	\caption{Face verification accuracy of models with different dimensions, depths, and parameters on CFP-FF, CFP-FP and VGG2-FP. 
% 	The top first three models lists the results of the CNN models, while mid-block presents the performance of pure ViTs architectures. The third block shows the results of ConViTs models.
}
	\label{tab:table_2}
	\vspace{-5pt}
\end{wraptable}
On the VGG-FP dataset, IPE increases performance with little margin on VGG2-FP from 57.1\% to 62.2\%. Overall, IPE improves average performance by 4.5\%. 
The introduction of convolutional FFN in FPVT improves the structural and local relationship between different parts of faces. Interestingly, CFNN adds significant performance gain on all datasets: LFW (3.8\%), CFP-FF (1.1\%), CFP-FP (4.6\%), Age-DB (3.3\%), CA-LFW (2.8\%), CP-LFW (3.1\%) and VGG2-FP (3.9\%). We also evaluate the influence of the FDR layer on FPVT performance (see in Table. \ref{tab:table_2} and Table. \ref{tab:table3}). While retaining the same training and implementation details, we replace the previous layer with our "+IPE+CFFN" and it gradually increases the performance on six datasets. %Although, the performance on CP-LFW is slightly decreased.
The introduction of the FDR layer in FPVT discriminates the features among identities that lead to performance improvement in six datasets. As mentioned in Table. \ref{tab:table_2} and Table. \ref{tab:table3}, FDR improves the accuracy on LFW from 86.7\% to 87.4\%, CFP-FF from 86.6\% to 87.4\%, CFP-FP from 61.5\% to 61.5\%, Age-DB from 68.9\% to 70.1\%, CA-LFW 72.9\% to 73.9\%, and VGG-FP 66\% to 66.1\%. However, the accuracy of CP-LFW slightly decreases from 62.1\% to 61.1\%.
The introduction of data augmentation and F-SRA improves linear computation and reduces the number of parameters. The final number of parameters of FPVT is decreased from $33.3$M to $28.8$M which is smaller than the recent pure ViTs, ConViTs, and CNNs. Further, we adopt online data augmentation by using some off-the-shelf techniques. As shown in Table. \ref{tab:table3}, the accuracy improvement is observed as: LFW (87.4\% to 91.4\%), CFP-FF (87.4\% to 90\%), CFP-FP (61.5\% to 71.8\%), Age-DB (70.1\% to 74.5\%), CA-LFW (73.9\% to 77.4\%), CP-LFW (61.6\% to 68.9\%) and VGG2-FP (66.1\% to 75.3\%). Overall, online augmentation significantly increases performance on all datasets. While F-SRA reduces the overall number of parameters parameters, it increases the accuracy on individual datasets such as on LFW (91.4\% to 92.0\%), CFP-FF (90\% to 90.3\%), CFP-FP (71.8\% to 73.3\%), Age-DB (74.5\% to 77\%).% and VGG2-FP (75.3\% to 76.3\%). %However, on two datasets the performance is a little bit decreased like on CA-LFW (77.4\% to 77.0\%), and CP-LFW (68.9\% to 67.8\%). 
\vspace{-5mm}
\section{Conclusion}
\vspace{-2mm}
\label{sec:future}
A Face Pyramid Vision Transformer (FPVT) is proposed for FR and verification tasks. Within the FPVT framework, a convolutional feed-forward network is used to  encode local structural relations among different facial parts and to maintain long range relations. To ensure parameters reduction, a Face-Spatial Reduction Attention layer is introduced in the encoder  that efficiently decreases the number of parameters. Additionally, a Face Dimensionality Reduction (FDR) layer is used to ensure facial feature map compactness.  The proposed FPVT is evaluated on seven  datasets and compared with ten SOTA methods. The experiments have exhibited the robustness of the proposed algorithm.

\bibliography{egbib}
\end{document}

\maketitle

\begin{wraptable}{R}{0.41\textwidth}
    \vspace{-40pt}
    \scalebox{0.50}{
        \input{tables/arch.tex}
    }
\caption{Calculated settings and the design principles follow the same rules of PVT \protect\cite{wang2021pyramid}. $e$ denotes MLP ratio, whereas, $r$ represents resolution, and $n$ denotes the number of heads.}
\label{tab:arch4}
%\vspace{-5pt}
\end{wraptable}
\section{Model Details}
The proposed FPVT parameters are described as follows: For the $i_{th}$ stage, $p_i$ is the patch-size, $c_i$ is the number of output channel, $l_i$ is the number of layers in encoder, $r_i$ is the reduction-ratio in F-SRA, $h_n$ is the number of heads, $e_i$ is the expanding-ratio of convolutional FFN.

\begin{comment}
\begin{table}[t]
\scalebox{0.75}{
	\begin{tabular}{p{1.3cm}<{\centering}|p{0.8cm}<{\centering}p{0.8cm}<{\centering}p{1.2cm}<{\centering}|p{0.8cm}<{\centering}p{0.8cm}<{\centering}p{1.2cm}<{\centering}}
		\hline
		%\thickhline
		\rowcolor{mygray} 
		& \multicolumn{3}{c|}{Original} & \multicolumn{3}{c}{Aligned}  \\ \cline{2-7} 
		\rowcolor{mygray} 
		\multirow{-2}{*}{Identities} & Male   & Female & Total   & Male   & Female & Total  \\  \hline \hline     
		People                       & 265    & 265    & 530     & 263    & 263    & 526    \\
		Image                        & 55,306 & 51,557 & 106,863 & 47,196 & 44,516 & 91,712 \\ \hline
	\end{tabular}
}
\caption{Statistics of the cleaned version are aligned $112\times112$ and contain fewer identities than the original facescrub database.}
\label{tab:FACESCRUB}
\end{table}
\end{comment}
Following the design principles of SwinT \cite{liu2021swin} and PyramidT \cite{wang2021pyramid}, we utilize the small number of output channels in shallow stages and focus the major computational resource  in the middle stages. To provide instances of FPVT, we present only one model of our method which is presented in Table.~\ref{tab:arch4}. The  number of parameters of FPVT is smaller than ResNet-18 \cite{he2016deep}, IR-18 \cite{deng2019arcface}, IR-SE-18 \cite{hu2018squeeze}. 

\section{Inference Speed}
We evaluate the inference speed of our proposed FPVT architecture, in order to present its feasibility under limited computational resources on real-time applications. We compare the FPVT speed with general ViT models on LFW dataset.The proposed FPVT provides a better recognition accuracy with the inference speed of general ViTs is 0.37s per image whereas our FPVT achieves 0.32s.

\bibliography{egbib}

%% file: tables/arch.tex
\begin{tabular}{c|c|c|c}
	\hline\thickhline
	\rowcolor{mygray} 
	Stages                     & Output Size                                                                & Layer Name            & OPVT                                            \\
	\hline
	\hline
	\multirow{2}{*}[-2.5ex]{1} & \multirow{2}{*}[-2.5ex]{\scalebox{1.3}{$\frac{H}{4}\times \frac{W}{4}$}}   & Patch Embedding       & \multicolumn{1}{c}{$P_1=7$;\ \ \ $C_1=64$}      \\
	%\cline{3-7}
	                           &                                                                            & {Transformer Encoder} &                                                 
	$\begin{bmatrix}
	\begin{array}{l}
	R_1=8 \\
	N_1=1 \\
	E_1=4 \\
	\end{array}
	\end{bmatrix} \times 2$  \\
	\hline
	\multirow{2}{*}[-2.5ex]{2} & \multirow{2}{*}[-2.5ex]{\scalebox{1.3}{$\frac{H}{8}\times \frac{W}{8}$}}   & Patch Embedding       & \multicolumn{1}{c}{$P_2=3$;\ \ \  $C_2=128$}    \\
	%\cline{3-7}
	                           &                                                                            & {Transformer Encoder} &                                                 
	$\begin{bmatrix}
	\begin{array}{l}
	R_2=4 \\
	N_2=2 \\
	E_2=4 \\
	\end{array}
	\end{bmatrix} \times 2$ \\
	\hline
	\multirow{2}{*}[-2.5ex]{3} & \multirow{2}{*}[-2.5ex]{\scalebox{1.3}{$\frac{H}{16}\times \frac{W}{16}$}} & Patch Embedding       & \multicolumn{1}{c}{$P_3=3$;\ \ \  $C_3=256$}    \\
	%\cline{3-7}
	                           &                                                                            & {Transformer Encoder} &                                                 
	$\begin{bmatrix}
	\begin{array}{l}
	R_3=2 \\
	N_3=4 \\
	E_3=4 \\
	\end{array}
	\end{bmatrix} \times 2$ \\
	\hline
	\multirow{2}{*}[-2.5ex]{4} & \multirow{2}{*}[-2.5ex]{\scalebox{1.3}{$\frac{H}{32}\times \frac{W}{32}$}} & Patch Embedding       & \multicolumn{1}{c}{$P_4=3$;\ \ \ $C_4\!=\!512$} \\
	%\cline{3-7}
	                           &                                                                            & {Transformer Encoder} &                                                 
	$\begin{bmatrix}
	\begin{array}{l}
	R_4=1 \\
	N_4=8 \\
	E_4=4 \\
	\end{array}
	\end{bmatrix} \times 2$ \\
	\hline\thickhline
\end{tabular}